\documentclass[sigconf]{acmart}

\AtBeginDocument{%
  \providecommand\BibTeX{{%
    \normalfont B\kern-0.5em{\scshape i\kern-0.25em b}\kern-0.8em\TeX}}}

\setcopyright{rightsretained}
\copyrightyear{2020}
\acmDOI{}

\acmConference[SUM '20]{Workshop SUM '20: State-based User Modelling, The Thirteenth ACM International Conference on Web Search and Data Mining (WSDM '20)}{February 03--07, 2020}{Houston, TX}
\acmPrice{}
\acmISBN{}

\defcitealias{ong}{[2015]}
\defcitealias{conley}{[2013]}
\defcitealias{murre}{[2015]}
\defcitealias{grutzik}{[2017]}
\defcitealias{dota2cp}{[2013]}
\defcitealias{agarwala}{[2014]}
\defcitealias{kalyanaraman}{[2014]}
\defcitealias{almeida}{[2017]}
\defcitealias{chen}{[2017]}
\defcitealias{sapienza}{[2018]}

\robustify{\centering}

\begin{document}

\title{Scalable Psychological Momentum Forecasting in Esports}

\author{Alfonso White, Daniela M. Romano}
\email{[a.white.16, d.romano]@ucl.ac.uk}
\affiliation{%
  \institution{University College London}
  \city{}
  \country{}
}

\renewcommand{\shortauthors}{White, Romano}

\begin{abstract}
  The world of competitive Esports and video gaming has seen and continues to experience steady growth in popularity and complexity. Correspondingly, more research on the topic is being published, ranging from social network analyses to the benchmarking of advanced artificial intelligence systems in playing against humans. In this paper, we present ongoing work on an intelligent agent recommendation engine that suggests actions to players in order to maximise success and enjoyment, both in the space of in-game choices, as well as decisions made around play session timing in the broader context. By leveraging temporal data and appropriate models, we show that a learned representation of player psychological momentum, and of tilt, can be used, in combination with player expertise, to achieve state-of-the-art performance in pre- and post-draft win prediction. Our progress toward fulfilling the potential for deriving optimal recommendations is documented.
\end{abstract}




\maketitle

\section{Introduction}

{\itshape League of Legends}, by Riot Games Inc., is one of the most popular video games worldwide, the most played of the multiplayer online battle arena (MOBA) genre. In August of 2019, there were an estimated 8 million peak concurrent daily players, with a total of more than 200 million active monthly players, and the game has consistently been among the most watched games on video platforms such as Twitch.tv and YouTube. It is considered one of the more difficult games to master, in part owing to its complexity, intensity and pace, imperfect information scenarios, snowballing effects, and depth of game knowledge required.

In particular, the number of characters to select from in the pre-match team draft phase create an enormous strategic search space, and, since the early days of MOBAs, recommendation engines have been used by a large number of players to help focus planning and ease personal learning \cite{conley}. In addition to the known aspects of player expertise in particular roles and compositional viability, there exists another predictor of in-game success, which, until now, has not been tapped for the potential of game plan recommendation: that of short-term performance deviation. This can be roughly subdivided into two forms: psychological momentum; usually a positive influence, or tilt; a negative influence similar to the notion of negative momentum. Psychological momentum is an important concept in sports science \cite{crust}, and can be defined as athletes' cognitive, affective and physiological disposition toward repeating the results of the previous event(s). It is closely related to the hot hand phenomenon, originating in basketball \cite{csapo}, where players may attempt more difficult shots, and have more success in them, when they are on a streak of points. It has been linked to flow states in elite sport \cite{swann}, and can occur in high performance work environments. {\itshape Tilt}, originating as a poker term, describes a suboptimal state of mind of which can occur after experiencing a significant loss \cite{palomaki}, whether a consequence of bad luck, having made a mistake, or a provocating exchange with an adversary. The term has since been adopted in the gaming community \cite{wei}, though, the same emotional mechanisms are observed in more universal scenarios, such as in {\itshape road rage}. Tilt in Esports and internet gaming is fairly common, with the general description, nonspecifically related to ``emotional breakdown and frustration, due to negative results following hard work'', occupying the top position on Urban Dictionary, with the example given citing {\itshape League of Legends}. We note that, while a player who is tilted is almost certainly experiencing negative momentum, one who is in a state of negative momentum may not necessarily be `on tilt'.

To account for these factors in state-based player recommendation, game design and analytics, we use various methods to accommodate participant's immediate historical performances within a win prediction model. We achieve a state-of-the-art classification accuracy of 72.1\% for {\itshape League of Legends} using a logistic regression, a 2.0\% improvement over previous work based on strategic player behaviour profile clustering \cite{ong}. We also implement a recurrent network that achieves a 0.5\% relative gain in pre-draft single player classification rate, with both models using an automatic logarithmic scaling that improves accuracy by up to 1.3\% for linear models and 2.8\% for neural networks. With machine learning, we are able to learn atypical, subtle and complex nonlinearities corresponding to short-term fluctuations in momentum, such as tilt onset, from a large, comprehensive dataset of player histories.


These models may be used to recommend better draft choices that synergise with the player's own ability more precisely, as well as with the team draft. For example, when significant positive momentum is detected, higher impact roles and characters that require a greater level of finesse or concentration may be suggested. By transferring pre-trained weights of the recurrent network submodules, we learn a first approximation to a pre-draft single player performance model conditioned to factor out players' baseline skill, estimating the effect of momentum alone. In the future work section, we describe our proposal to use this player state representation as input to a tilt recognition model, and a reinforcement learning agent that can accurately coach the player in tilt management, using sympathetic between-match notifications. These can contain suggestions for when and how to take optimal breaks, a tilt reduction strategy that can improve mentality and performance in {\itshape League of Legends} \cite{kou}.

\subsection{Game Description}

\begin{figure}[!ht]
\centering
\includegraphics[height=62.5mm]{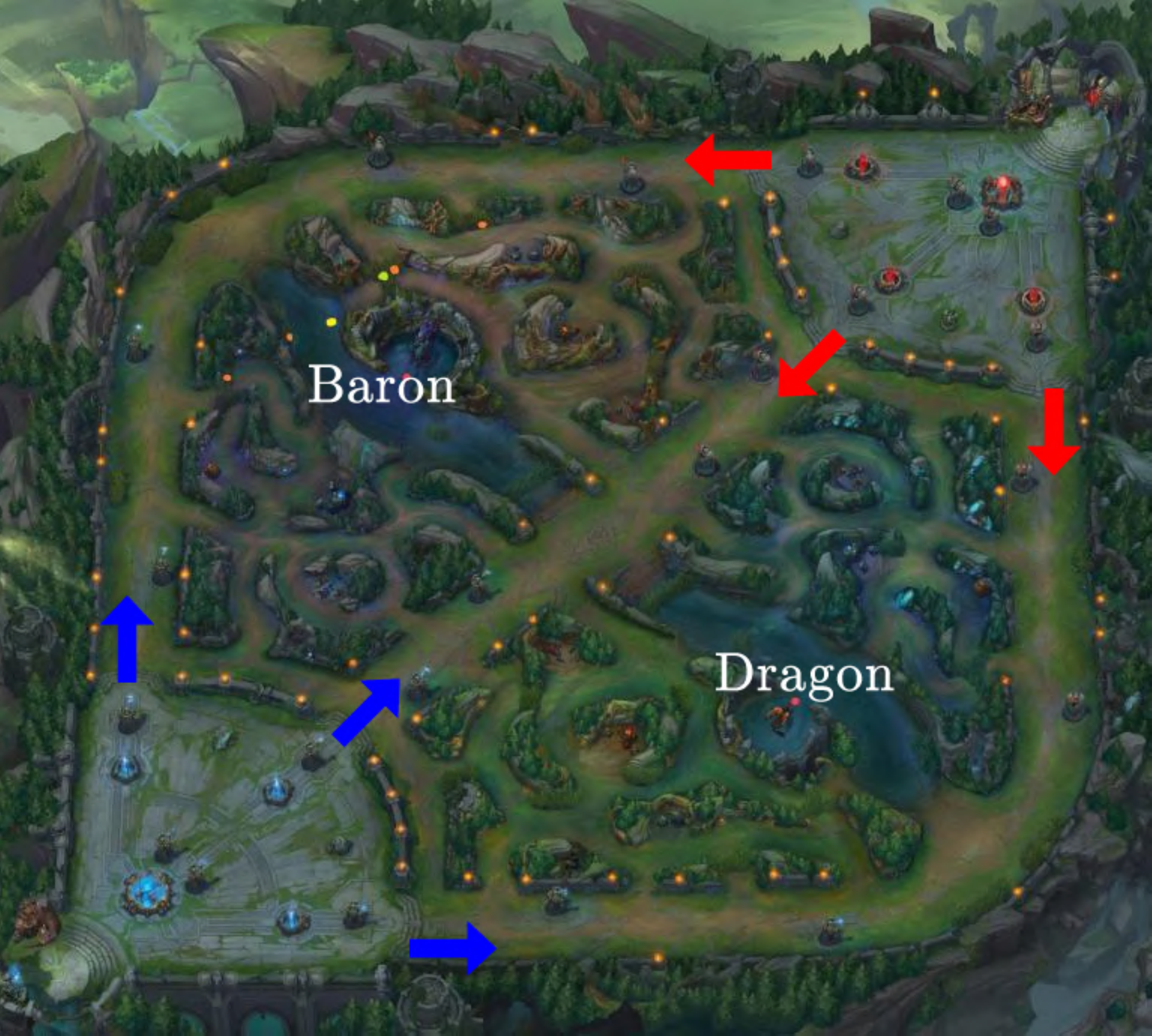}
\caption{Map of Summoner's Rift.}\label{fig:srMap}
\end{figure}

In {\itshape League of Legends}, the player assumes the role of a summoner, who summons one of 148 unique champions onto The Rift, the environment in which the game is played (fig. \ref{fig:srMap}). Two teams of five players, spawning in opposite corners of the map, are pitted against each other in a race to build up strength through earned gold and items, strategically advance upon the opponent's base, and ultimately win by destroying the nexus. Each champion has their own set of abilities, playstyle, and strategic position, though these can be approximately categorised into a number of overlapping classes such as {\itshape Mages} (spell casters), {\itshape Marksmen}, {\itshape Tanks}, {\itshape Fighters}, and {\itshape Supports}. The strengths of each character vary between physical locations on the map, and across the temporal game phases of each match, which in total last an average of 32 minutes. There are five main positions, or roles, which players occupy: {\itshape Top}, {\itshape Jungle}, {\itshape Mid}, {\itshape Bottom}, and {\itshape Support}, and a player can optionally select a role for the matchmaking algorithm to prioritise, in addition to queueing for a match at the player's skill level (Elo rating). The number of champions and therefore the number of possible combinations of interactions between teams, make each game, though played in an identical setting, unique in nature and line. This creates a diverse set of learning scenarios in the zone of proximal development.

\section{Related Work}

The field of win prediction in competitive gaming and Esports is one of active research, both in the pre-game setting, and in real-time prediction. The first work on draft recommendation in a MOBA game was DotA2CP \citetalias{dota2cp}, for one of the first major titles, \emph{DotA 2} (successor to \emph{Defence of the Ancients}, and the most often studied MOBA), reportedly achieving an accuracy of 63\% using hero picks alone. Conley and Perry \citetalias{conley} built upon this work to achieve 69.8\% accuracy with logistic regression and 70\% accuracy with k-nearest neighbours (k-NN), still using only the presence of heroes on either team as training data. They also created a pick recommendation engine using a greedy algorithm to add heroes to a team incrementally based on updated win probability. Agarwala and Pearce \citetalias{agarwala} also used logistic regression, but added a prior principal component analysis (PCA) step in order to study team composition. This did not increase the predictive accuracy on the match history dataset, however, it did improve the pick recommendations given, as it caused the model to capture more about interactions between heroes, rather than assuming that players have already chosen a balanced combination (out of distribution generalisation). Kalyanaraman \citetalias{kalyanaraman} was the first to introduce the hero roles as a model feature, and used a combination of a genetic algorithm and logistic regression to achieve 74.1\% accuracy on a dataset of high skill rank matches in {\itshape DotA 2}. Almeida et al. \citetalias{almeida} used Naive Bayes to achieve an accuracy of 76.3\%. Sapienza et al. \citetalias{sapienza} performed an analysis using neural networks to recommend teammates for advancing in \emph{DotA 2}.


Ong et al. \citetalias{ong} used k-means clustering of strategic player behaviours and a support vector classifier (SVC) to achieve 70.4\% in post-draft {\itshape League of Legends} win prediction. Chen et al. \citetalias{chen} examined player skill in \emph{League}, finding that player's base skill, their chosen champion's base skill, and the player's skill on that champion are the top three components, and were able to score 60.24\% using logistic regression (LR). To the best of our knowledge, our work is the first to succeed in employing players' immediate history to improve a win prediction model; previous work by Grutzik et al. \citetalias{grutzik}, on Esports win prediction in {\itshape DotA 2}, made an attempt at this, using neural networks and rolling statistics for the last 10 professional matches. Other work has used hierarchical attention-based networks to recommend purchasable items in the mobile MOBA {\itshape King of Glory} \cite{yao}, and there is much research on recurrent models for in-game prediction \cite{lan}. In terms of in-production recommendation and coaching tools for {\itshape League}, the most popular is {\itshape \href{https://blitz.gg}{Blitz}}, with over 1.5 million users, providing matchup-based champion suggestions, optimal pre-match runes, item build paths, and informative post-game analysis. The field of reinforcement learning has also made much progress in solving MOBA games as a stepping stone toward solving artificial general intelligence (AGI) \cite{zhang}, as it provides an example of a complex, co-operative, real-time task with sparse and delayed reward signals. In April of 2019, Open AI Five defeated the world champion \emph{DotA 2} team, OG \cite{openai}. Whereas other tilt detection methods have used peripheral equipment to estimate affective states \cite{wei}, we use a minimal amount of information to infer effects through user interactions and performance statistics alone, scaling to production with zero user requirements.


\section{Dataset}

A number of sources are combined in order to efficiently obtain a dataset. The Riot Games developer API is used to crawl player match histories, searching for games which occurred recently, contain unseen players (at least from the recent past), and such that the skill distribution is sampled uniformly. Once a valid match is found, participant profile summaries are loaded from \href{https://op.gg}{op.gg}, which contain on-champion proficiency statistics, season totals, and performances for up to the last 20 games. In addition, global and regional averages for each champion and common matchups (pairs of champions often found competing in the same lane) are loaded from \href{https://champion.gg}{champion.gg}, and \href{https://op.gg}{op.gg}, and updated daily. This is to keep up with game updates, which are released approximately once a fortnight, and can have a dramatic impact on the strategic metagame.

In total, 87,743 valid samples were collected starting two weeks after the beginning of season 9, from February 5\textsuperscript{th} to September 20\textsuperscript{th} of 2019. 86.4\% were from Solo/Duo queue, 13.6\% from Flex queue, and in total contained 517,269 unique summoners. 70,194 matches were used in training within five 5-fold stratified cross validations, 7,743 in a validation set, and 10,000 in testing. The corresponding 701,940 individual match histories are used for training single player pre-draft models within the same folds.

\section{Methodology}

\subsection{Feature Engineering}

While our network model is powerful enough to encode an appropriate distributed representation from just the raw data, in practice, training this kind of model is difficult, partly due to the amount of data required, a result of the combinatorial explosion. Instead, we perform some preliminary feature engineering to reduce the complexity. For example, one feature, present for each role, is the global average matchup win rate for the specific pair of champions. The full feature set used is given in appendix \ref{sec:allfeats}. Features are standardised to zero median and unit interquartile range.

\subsubsection{Experimental momentum retention representation}
\begin{figure}[!ht]
\centering
\includegraphics[width=75mm]{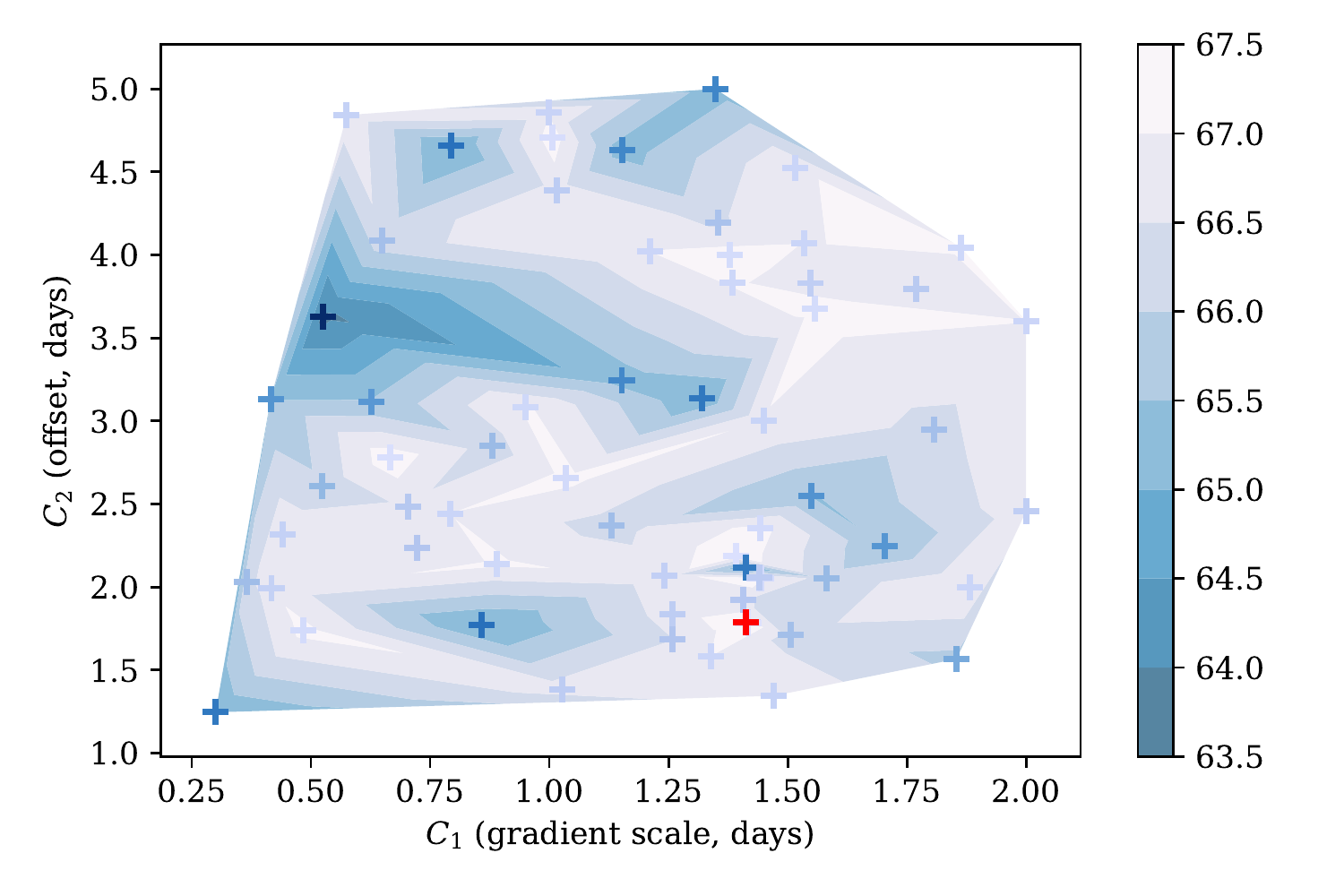}
\caption{Bayesian optimisation of momentum constants.}
This graph maps the performance of parameter pairs for the Ebbinghaus momentum representation. $C_1$ controls the rate at which the impact of past events decays exponentially, while $C_2$ adds a constant base rate. Skill retention timeframes may also be relevant. Accuracy of the resulting feature set was assessed using a training set of 30,000 samples, and 10,000 for testing; improvements were consistent when using the full training dataset. The peak value of 67.40 is shown in red.\label{fig:mtmHypOpt}
\end{figure}

To attempt to capture the time dependent effects of the onset or dissipation of shorter term performance deviation, we investigate the use of a feature representation based on an exponential model of memory retention \cite{murre}\cite{ebbinghaus}. Effects of recent events on performance are approximated as random deviations that dissipate with time exponentially. These features are gathered multiple times, for the last 1, 2, 4, 8, and 20 matches. We also normalise by the match duration (rolling statistics are normalised only by match duration ($z$ values), and gathered for the last $t\leq20$ recent matches).
  \begin{align}
    z &= v \times \frac{\bar{d}}{d} \\
    x &= \frac{z}{1 + \text{exp}((C_{1} \times (\mathfrak{t}_{\text{now}} - \mathfrak{t})) - C_{2})}
  \end{align}
  where \\
  \hspace*{5px} $v$ is one of the recent history values, prior to normalisation, \\
  \hspace*{5px} $d$ is the recent game duration ($\bar{d}$ is the mean duration of 32 mins), \\
  \hspace*{5px} $\mathfrak{t}$ is the match end timestamp in days ($\mathfrak{t}_{\text{now}}$ is the current time), \\
  \hspace*{5px} $x$ is the final, Ebbinghaus-normalised version of $v$, and \\
  \hspace*{5px} $C_{1} = 1.41$ and $C_{2} = 1.79$ (days) are learned by a Bayesian optimisation of Gaussian process upper confidence bound (fig. \ref{fig:mtmHypOpt}). Normalising by past match duration increases accuracy by 0.2\% for the logistic regression. 


\subsubsection{Rolling statistics medium-term momentum representation}

Our primary momentum features, used for all models, are the performance summaries for the last $t\leq20$ matches played for each player. These contain the player's performance scores for the games, as well as the differences between these scores and the player's average for the champion (for the current season, or over the last two seasons if the sample size is small). $v^{\text{(momentum)}}_{t} = v_{t}-\bar{v}_{(\text{season})}$.

\subsubsection{Automatic Logarithmic Scaling}

\begin{figure}[!ht]
\centering
\includegraphics[width=\linewidth, trim={0 0 0 0}, clip]{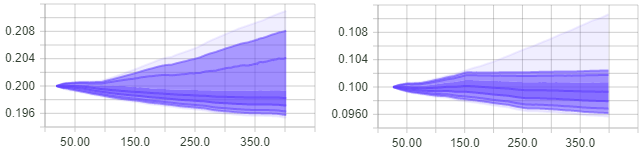}
\caption{\centering Distribution of $\alpha$ values in the AutoLog layer for scalar features, over training, before and after adjustment.}
Samples are taken for each of 400 batches of 256 data points.\label{fig:alphaDist}
\end{figure}

The scalar inputs to our models are close to normally distributed, however, many contain long tails in the positive direction. This is due to the nature of their generation: scores are bound to be above zero, not bound in the positive direction, snowball, and, the duration of the game is also unbound. To counteract this, a logarithmic transformation can be applied, curbing the skew, however, each input's skew is distinct, meaning a choice of scale factor and offset for each input is needed; a difficult and impractical task for the number of features used, and with frequent game updates. To solve this we propose and implement a logarithmic scaling layer, which transforms scalar inputs prior to modelling. Initial seed factor and offset parameters $\alpha, \beta$ and $\gamma$ are manually specified to minimise the spread from the initial locations in training (fig. \ref{fig:alphaDist}), then optimised using Bayesian hyperparameter optimisation beginning in a small enclosing region, which is updated if necessary. For an input data point $x \in \mathbb{R}^m$,
\begin{align}
\text{AutoLog}(x) &= \alpha \circ x + \log(\beta \circ (\gamma + \max(0, x - \min(X_{\text{train}})))) \nonumber \\
\min(X_{\text{train}}) \in \mathbb{R}^m&\;\;\; \text{ are the approx. min. values for the features} \nonumber \\
\alpha_{\text{init}} &= \mathbf{0.1} \nonumber \\
\beta_{\text{init}} &= \mathbf{0.4},\; \forall i . \beta_i > 0 \nonumber \\
\gamma_{\text{init}} &= \mathbf{0.1},\; \forall i . \gamma_i \geq 10^{-4} \nonumber \\
\circ&\;\;\; \text{ is }\text{the Hadamard or entrywise product} \nonumber
\end{align}
This significantly improves accuracy for the neural network model, particularly when a separate set of initial seed parameters are found for the recurrent inputs (6 values in total). Average final seed $\alpha, \beta$ \& $\gamma$ for the scalar inputs are 0.054, 0.372, and 0.006, and for the recurrent inputs, 0.123, 0.211, 0.073. For the logistic regression, a set of $\alpha, \beta$ and $\gamma$ shared for all features obtains a better performance, and, the low time complexity means a bayesian optimisation can be used to find the optimum, however, feature-specific AutoLog parameters learned from the neural network training perform 0.1\% better than those learned from a logistic regression optimised by gradient descent, indicating that transfer learning may be possible in a similar scenario. In both cases the use of an automatically configured logarithmic scaling accelerated the process of finding desirable optima, with the inclusion of the $\alpha$ parameter increasing accuracy by 0.15\% for the neural network and 0.05\% for the logistic regression. In general, this layer may reduce the amount of time needed to find the optimum of logarithmic scaling parameters for unstable or mixed-distribution inputs to a high complexity model.

\subsubsection{Feature Selection}

The final feature set (appendix \ref{sec:allfeats}) is selected from a bank of features which is approximately twice as large, using a sequential forward floating selection (SFFS) to optimise cross validation score. This is made computationally feasible by using logistic regression as the model, mean-averaging expert momentum features for each role, mean-averaging features across the members of each team, and taking the difference between teams.

\subsection{Model Architectures}

\subsubsection{Linear Model}
Our top performing architecture for post-draft win prediction is a logistic regression with $\ell_2$-regularised weights optimised by L-BFGS. This type of linear model directly optimises the difference in predictive distribution and is resilient to noise in large feature and data sets, making it suitable for our task. We found it to outperform many other linear and nonlinear models of varying hypothesis space complexity. 3 runs of 3-fold stratified cross validation are used internally on the 70,194 training samples to choose an appropriate inverse regularisation parameter $C$ from the range [$e^{-2}$, $e^{4}$]. Due to the limited ability for continuous and complex temporal dependencies to be learned through linear models and feature engineering alone, we also experiment with a suitable nonlinear algorithm that can be applied to our dataset.


\begin{figure}[!ht]
\centering
\includegraphics[width=\linewidth]{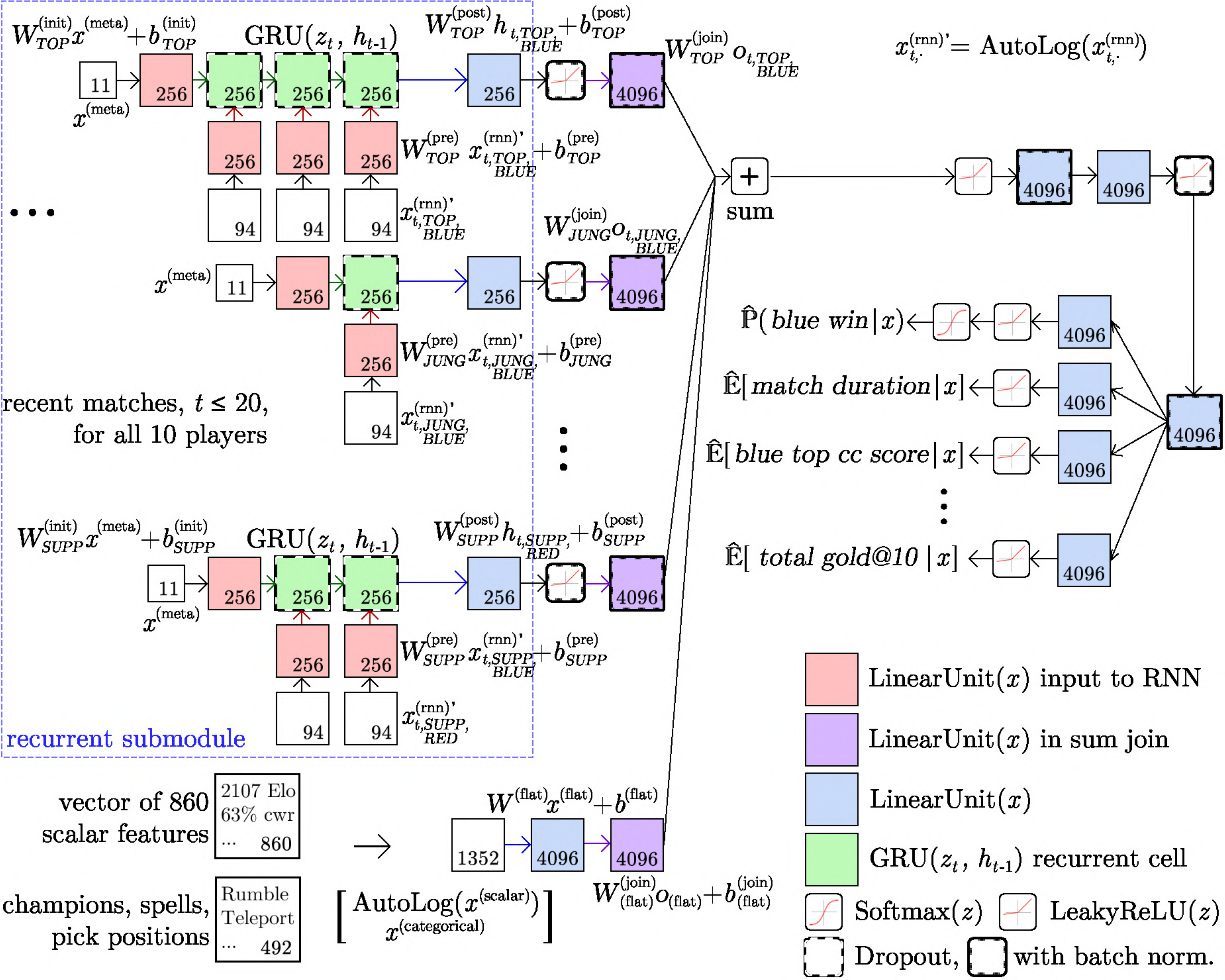}
\caption{Win prediction multi-task recurrent network.}
$x^{(\text{meta})}$ consists of the player Elo and region, which are also included in $x^{(\text{flat})}$. $x^{(\text{rnn})}_{t, \cdot}$ contains the time since the recent match occurred, time of day, duration, overlapping class membership-encoded champion, and statistics. Linear layers surrounding recurrent cells are role-convolutional.\label{fig:networkDiagram}
\end{figure}

\begin{table}[h!]
\centering
\caption{\centering Multi-task learning targets for post-draft network.}\label{tab:modelOutputs}
\begin{tabular}{ll}
 \toprule
 Target (\# instances) & Description \\
 \midrule
  Blue side win (1) & Single classification target  \\
  Match duration (1) & Length of the game \\
  Crowd control (10) & Type-normalised cc score, per player \\
  Vision score (10) & \# wards placed or destroyed, per player \\
  CS@10 (10) & Creep score at 10 mins, per player \\
  Total gold@10 (1) & The sum of all players' gold at 10 mins \\
  \bottomrule
\end{tabular}
\end{table}

\subsubsection{Neural network} Our post-draft neural network is as follows (fig. \ref{fig:networkDiagram}). Two input submodules, a recent past recurrent component (with a GRU cell to prevent vanishing gradient), and the flat inputs, are joined by sum operations prior to four hidden layers of 4096 units each. A recurrent sequence is used for each of the ten players' recent matches, sharing the same recurrent cell weights, however, a linear layer with 256 units is used before the recurrent cell, with weights shared only for the same roles (5 total weight matrices), transforming each player's performance into a role-independent activation. Another fully connected layer of 256 units with role-shared weights is used after the recurrent layers, accounting for role-specific momentum dependencies, before the fully connected layer for the join, which uses 4096 units. An initial state vector for the recurrent cell is also computed using a linear layer, from categorical region and scalar Elo metadata - this gives the recurrent module a point of reference to differentiate skill from momentum, especially when $t$ is low; performance is measured relative to the baseline skill for the region, Elo, and role. The 860 scalar inputs are combined with the 10 champions, draft pick ordering, and summoner spells for each player, before a single linear layer with 4096 units for the join. The 10 champion choices are summed for each team, reducing the number of required inputs from 1480 to 296, while maintaining accuracy. Dropout and batch normalisation are used at many points in the graph to prevent overfitting; keep probabilities are 0.55 between recurrent layers and 0.67 between non-recurrent layers. LeakyReLU activations are used to control gradients without complete deactivation. AMSGrad is a stochastic gradient descent variant which adds a fraction of the maximum of past squared gradients to the current update vector, reducing rare informative minibatch diminishing that occurs with the exponentially decaying averages of typical variants (i.e., Adam). We use it to optimise a sum of the win classification log-loss and the mean of regression $\ell_2$ losses, with a ratio of 0.01 (the scale of the $\ell_2$ sum is much larger; this value equalises scale and slightly prioritises the classification loss). Each regression output (min-max rescaled to between 0 and 1) is weighted equally. A learning rate of $3.5 \times 10^{-5}$ is used, with $\beta_1 = 0.9$ and $\beta_2 = 0.99$. Architectural importances are given in table \ref{tab:modelImprovements}, and the multiple simultaneous learning tasks in table \ref{tab:modelOutputs}.

\begin{table}[h!]
\centering
\caption{\centering Post-draft network architecture importances.}\label{tab:modelImprovements}
\begin{tabular}{lc}
 \toprule
 Improvement & \% Gain \\
 \midrule
  AutoLog layer (vs. shared $\alpha,\beta,\gamma$/untransformed) & 0.904/2.771 \\
  Recurrent structure (RNN) (vs. Rolling/Ebbinghaus) & 1.473/2.621 \\
  AMSGrad \cite{reddi} (vs. SGD/Adam) & 0.643/0.587 \\
  Multi-Task Learning (MTL) (vs. Win only) & 0.327 \\
  Metadata RNN initial state $Wx + b$ (vs. $b$ alone) & 0.209 \\
  Gated Recurrent Unit (GRU) cell (vs. $\tanh$) & 0.158 \\
  Role convolutions (vs. fully shared linear layers) & 0.073 \\
  Dropout \& Batch norm. on non-recurrent layers & 0.064 \\
 \bottomrule
\end{tabular}
\end{table}

When estimating pre-draft solo win probability, an alternative structure is employed to maximise predictive accuracy by extracting a useful momentum embedding: first, we learn the single post-draft classification task, abridging the four layers of 4096 units, and, the output dimension of the post-RNN layer is set to 32 units, with no activation or dropout (only batch normalisation). All scalar inputs are concatenated with the recurrent modules directly prior to the output layer; the logistic loss is minimised using Adam (learning rate $4.0 \times 10^{-4}$), on the recurrent submodule output, the rolling features, the probability given by the previous logistic regression (trained with L-BFGS), and the remaining scalar features (categoricals are not included). We then generate role-specific momentum embeddings using the trained recurrent submodule, reduce the dimensionality from 32 to 2 using principal component analysis, and include these two components (e.g., skill and momentum) as player inputs to the original logistic regression trained with L-BFGS. %

\section{Experiments}

\subsection{Model Evaluation}

Our techniques are compared in table \ref{tab:modelComparison}. The representational capacity of deep networks makes them the most suitable choice for momentum estimation, though the logistic regression outperforms in the multiplayer prediction task, with rolling statistics preferred over the Ebbinghaus features. Logistic regression accuracy plateaued by $\sim$70,000 data points. TensorFlow and four Nvidia Tesla v100s were used for implementation. With gains in normalisation, escaping local minima, maintaining gradient flow, and data (sec. \ref{sec:dataRes}), higher scores are possible. Nondeterminism and noise are key.

\begin{table}[h!]
\centering
\caption{\centering Learning algorithm comparison. \normalfont{This may also show the change in predictive game factors since earlier studies.}}\label{tab:modelComparison}
\begin{tabular}{lccc}
 \toprule
 Algorithm & Train $n$ & Test \% & Train \% \\
 \midrule
  AutoLog+Rolling\textsuperscript{+(momentum)}+LR & 70,194 & 72.1 & 73.6 \\
  AutoLog+Rolling+LR & 70,194 & 72.0 & 73.5 \\
  AutoLog+LR & 70,194 & 71.8 & 72.8 \\
  AutoLog+MTL+RNN & 70,194 & 71.1 & 71.6 \\
  Log\textsubscript{init}+Rolling\textsuperscript{+(momentum)}+LR & 70,194 & 70.8 & 73.1 \\
  k-means+SVC \cite{ong} & 117,000 & 70.4 & 74.8 \\
  k-means+LR \cite{ong} & 117,000 & 68.8 & 74.8 \\
  Rolling\textsuperscript{+(momentum)}+LR & 70,194 & 68.8 & 71.7 \\
  LR (baseline) & 70,194 & 68.3 & 71.4 \\
  LR \cite{chen} & 208,091 & 60.24 & - \\
 \midrule
 Pre-draft Teams & & & \\
 \midrule
  AutoLog+Rolling\textsuperscript{+(momentum)}+LR & 70,194 & 65.7 & 66.8 \\
  AutoLog+Rolling+LR & 70,194 & 65.6 & 66.7 \\
  AutoLog+LR & 70,194 & 65.1 & 66.0 \\
  AutoLog+MTL+RNN & 70,194 & 64.4 & 64.5 \\
  Log\textsubscript{init}+Rolling\textsuperscript{+(momentum)}+LR & 70,194 & 62.9 & 66.2 \\
  Rolling\textsuperscript{+(momentum)}+LR & 70,194 & 62.7 & 65.8 \\
  LR (baseline) & 70,194 & 62.3 & 65.4 \\
  MTL+RNN & 70,194 & 61.6 & 68.9  \\
  LR \cite{chen} & 208,091 & 56.75 & - \\
 \midrule
 Pre-draft Solo (in-queue) & & & \\
 \midrule
  AutoLog+Rolling\textsuperscript{+(momentum)}+RNN & 701,940 & 54.30 & 54.36 \\
  AutoLog+Rolling\textsuperscript{+(momentum)}+LR & 701,940 & 54.28 & 54.34 \\
  AutoLog+Rolling+LR & 701,940 & 54.03 & 54.12 \\
  AutoLog+LR & 701,940 & 53.59 & 53.71 \\
  LR (baseline) & 701,940 & 53.38 & 53.49 \\
  AutoLog+MTL-TL+RNN & 701,940 & 52.48 & 52.70 \\
  AutoLog+RNN & 701,940 & 52.38 & 52.72 \\
 \bottomrule
\end{tabular}
\end{table}
\begin{figure}[!ht]
\centering
\includegraphics[height=35mm, trim={7.3cm 1.3cm 0 1.5cm}, clip]{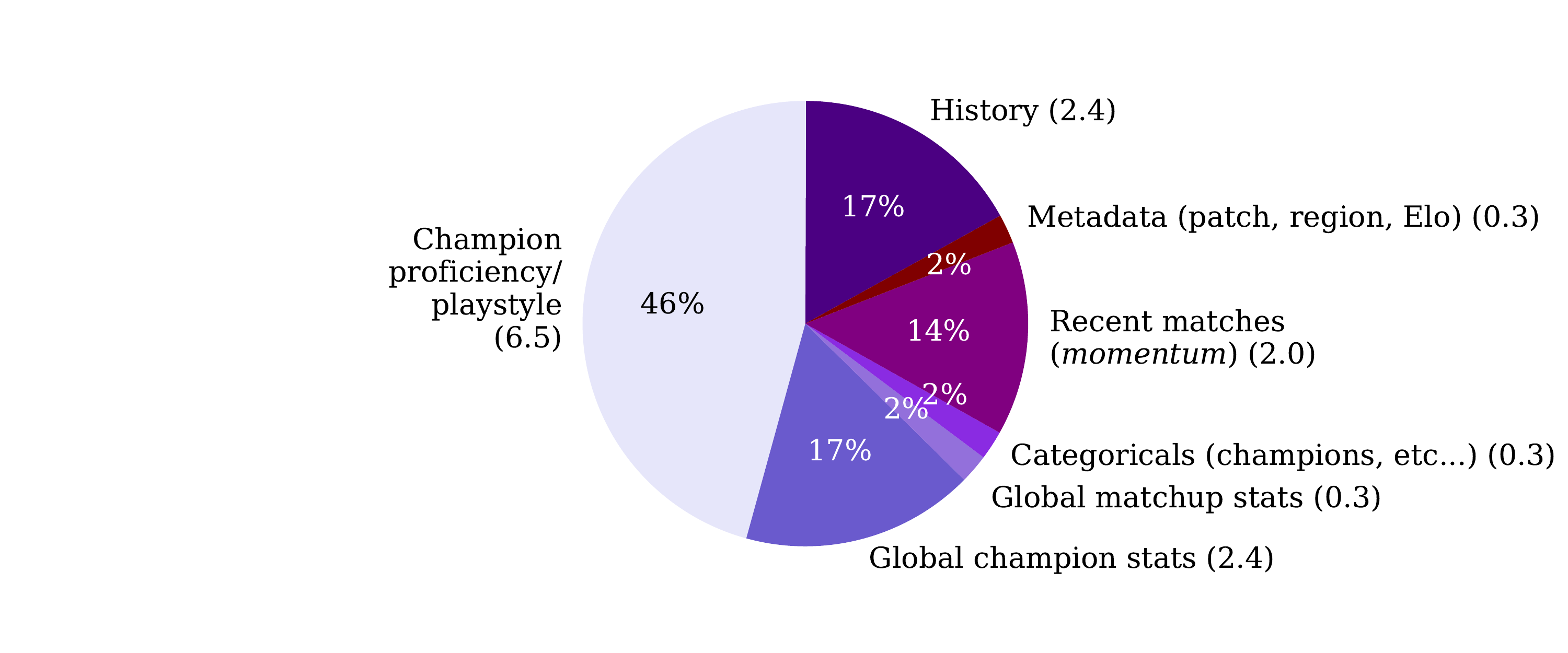}
\caption{Approximate Feature Group Importances.}
Drop in accuracy when the group is omitted (in brackets); relative importance is illustrated as a fraction of the total importance.\label{fig:featImps}
\end{figure}

\subsection{League of Legends Game Factors}

To study the factors that determine the outcome of a match, we observe the contribution in neural network accuracy when including independent feature groups (fig. \ref{fig:featImps}). Proficiencies are the most significant factor post-draft. The choice of champion may be influenced by momentum, though this is contextual, depending both on the team composition and the player's intentions. Of the information available prior to champion select (in-queue), the 2\textsuperscript{nd} largest contribution, around 40\%, comes from the most recent matches.


\subsection{Influence and Momentum Models}

\begin{figure}[!ht]
\centering
\includegraphics[height=38mm, trim={0.2cm 7cm 15.3cm 0cm}, clip]{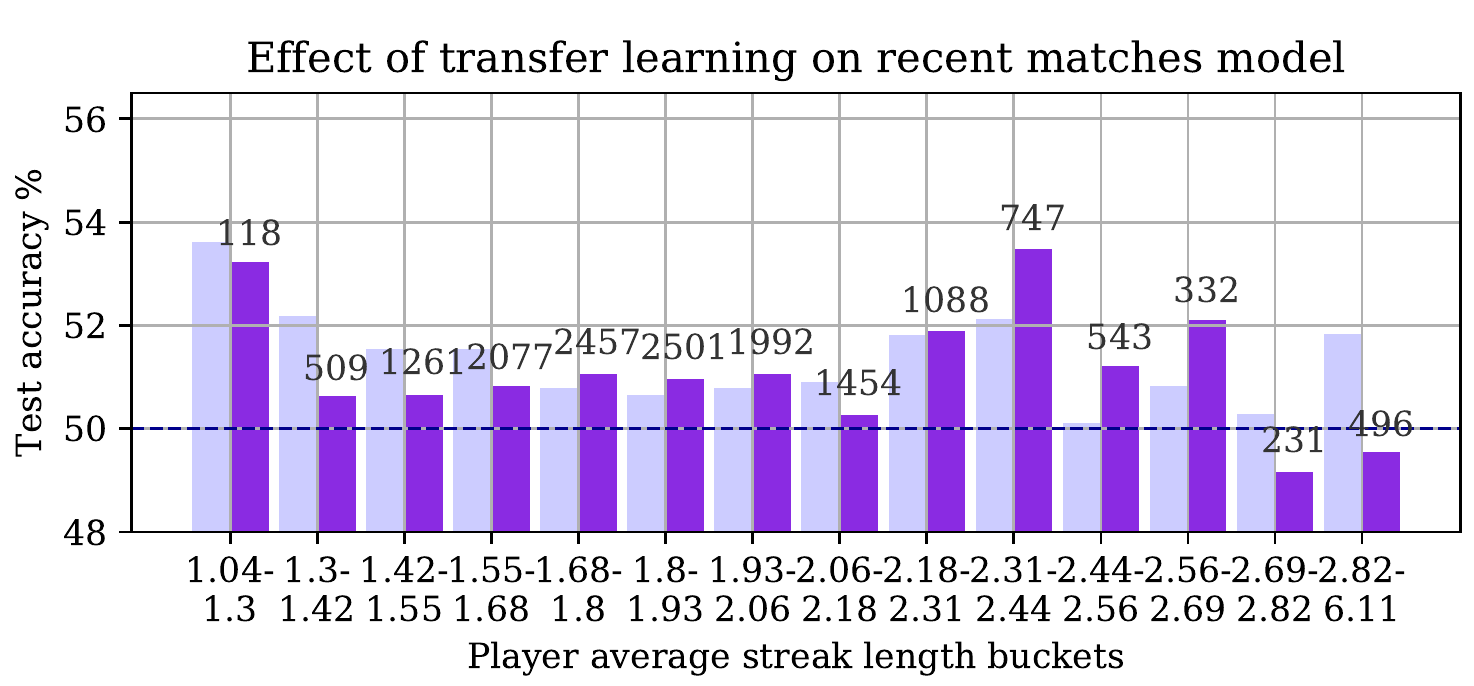}
\caption{\centering Accuracy for representative groups of players with similar streakiness tendency (either win or loss).}
We examine players with 2 or 3 occurrences in test dataset, and $n\geq15$ historical recent matches belonging to a complete streak, of known length, and $n\geq15$ belonging to a complete session). Allowing outliers at the two extremes, the transfer-learned model (violet) shows a slight positive relationship. Annotations are the number of data points (players) used to compute each bar.\label{fig:momentumWins}
\end{figure}

\begin{figure}[!ht]
\includegraphics[height=31.4mm, trim={0.3cm 0.3cm 0.15cm 0.3cm}, clip]{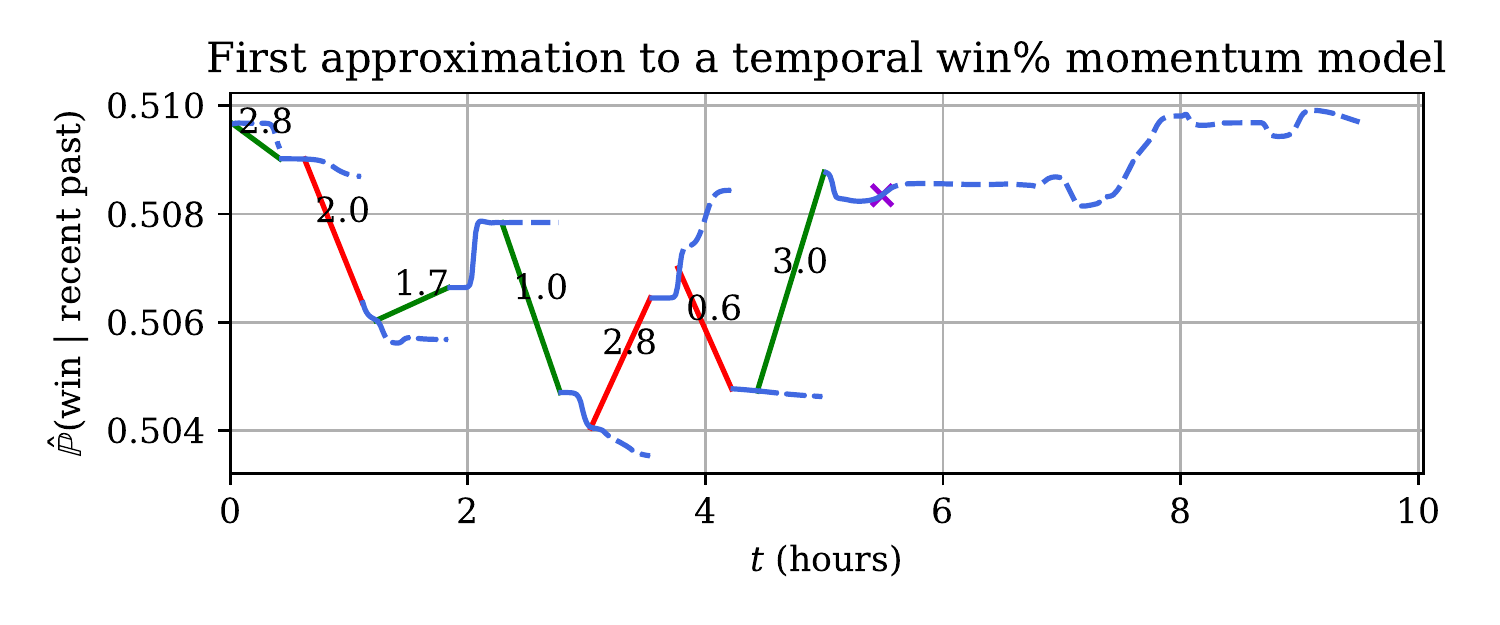}
\includegraphics[height=31mm, trim={0.3cm 0.3cm 0.15cm 0.3cm}, clip]{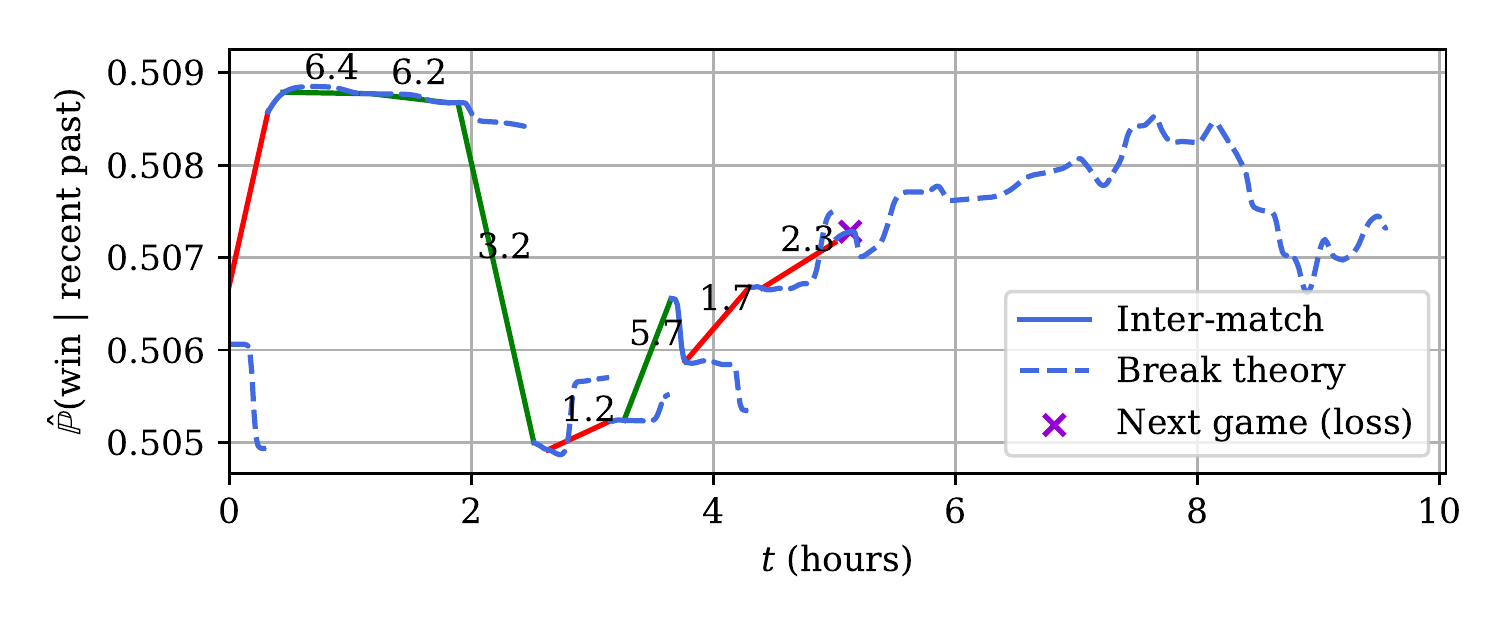}
\caption{Two players' momentum estimates over time.}
Past wins are represented by green lines, losses by red, and each match is annotated with post-game KDA ratio (kills + assists $\div$ deaths); a value of 2.0 is considered average. Tilt is recognised as a point where a post-match break increases win probability after around 20 minutes, though actively prompting the player to temporarily disengage, and providing an objective reason to (fig. \ref{fig:flowApp}), is likely to accelerate effects and aid tilt recovery \cite{kou}.\label{fig:mtmTime}
\end{figure}

We experiment with two single player pre-draft models, the RNN \emph{win\% influence}, and the \emph{win\% momentum} (recent effects, independent of baseline skill). Both are composed of the pretrained recurrent submodule from the full post-draft model, subsequently using two 512-unit layers. The influence model also uses a 512-unit layer to sum-join scalar history features, and consistently achieves a slightly higher accuracy than the same model without pretraining (52.48 over 52.38\%). Win\% momentum is compared with average player streak size for players with 2 or 3 occurrences in the test dataset (fig. \ref{fig:momentumWins}), and short-term momentum conditioning is observed (fig. \ref{fig:mtmTime}). While the model accuracy is below that of a logistic regression, it has learned a short-term temporal structure which the logistic regression is unable to fully capture. The nonlinearities correspond with the expected hidden latent construct, in that they predict based on a player's typical streakiness tendency, and show temporal behaviours consistent with the hypothesis. By quasi-marginalising player history and skill via the use of the pretrained recurrent submodule, we achieve an intuitive and unbiased metric.

\section{Discussion}

Though the signal found is faint, we are able to account for momentum that occurs over the medium term of the last $t\leq20$ matches within a post-draft win prediction model, increasing the accuracy by 0.1-0.3\%, and, a recurrent network architecture is able to forecast short-term nonlinear fluctuations, increasing pre-draft solo accuracy by 0.02\%; a relative increase of 0.5\%. Significant improvements are possible, in pre- and post-draft settings, by reducing the noise-induced difficulty of capturing the underlying function, and with data (\ref{sec:dataRes}). The small effect size of transferring momentum embeddings from the recurrent network to the logistic regression indicates that the underlying structure may be hidden in the distributed representation, or that the gradient descent training is not able to sufficiently pick up on the subtleties. By including existing features in the encoder training, we attempt to learn a shorter-term representation which factors out the medium-term baseline (information already included in the rolling statistics). The form by which the model approximates the underlying function, and the methods needed to capture patterns, are valuable to study as they can contain potentially transferable information. Multi-task learning helps the learning process via inductive transfer from the task of predicting in-game statistics. Medium-term pre-draft momentum was significant; a 20\% relative gain in accuracy for the solo model. Up-to-date performance data may also allow a more accurate estimate by measuring skill relative to changes in game and metagame structure due to game updates. Regarding why tilt onset occurs, one reason may be that humans are blind to integration noise when accounting for multiple discordant sources of cognitive information. This noise blindness occurs even at lower difficulties, and is consistent across time; effects remain even when evidence of overconfidence in choices is shown \cite{herce}.

\section{Conclusions}

Overall, our analysis and experiments show that it is possible to model the phenomena of psychological momentum and tilt in their context-sensitive impact on the outcomes of competitive games. While the system introduced is designed for {\itshape League of Legends}, it can be directly applied to other MOBAs, to other genres, and, with adaptations, other activities outside of gaming (see sec. \ref{sec:otherAreas}). The probabilities returned intuitively reflected subjective predictions.

\subsection{Limitations}

\subsubsection{Data Resolution}\label{sec:dataRes}

Throughout our experiments, the dataset used, assembled from various high bandwidth, relatively open sources, has been sufficient to show that our methodology is feasible. However, the resolution of this data may be a limiting factor. Elo information for particular players was only requested after the target match ended, meaning that differences in skill rating prior to the match were not obtained, and thus only the average for all 10 participants could be used (while Elo may account for momentum, the stochasticity of the match result obscures this). Due to the summarised form of performance histories, the data we have for analysing nuances of tilt (and momentum) is relatively low resolution. Temporal in-game data may be highly valuable; post-match summaries cannot distinguish effects that have appeared over the course of a game from those which have dissipated.

\subsection{Future Work}

Here we describe our ongoing and future efforts in creating useful, momentum-sensitive recommendation systems.

\subsubsection{Draft Pick Recommendations}

Initial experiments using a greedy algorithm to select the next champion based on the increased win probability have been promising, however, the application of unbiased methods \cite{agarwala}, is a top priority. Field testing will be used to verify momentum-utilising suggestions.

\subsubsection{Application to DotA 2}

While player skill and other player factors are known to be less game-deciding in {\itshape DotA} \cite{chen}, momentum and tilt are still present, and the evaluation of our system on a more widely studied title will be informative.

\subsubsection{Flow App}

\begin{figure}[!ht]
\centering
\includegraphics[height=37mm]{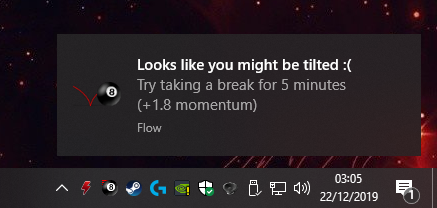}
\caption{Example \emph{Flow} notification.}\label{fig:flowApp}
\end{figure}


{\itshape Flow} (fig. \ref{fig:flowApp}), is a small desktop applet that subtly guides players to be more successful using illuminating notifications and tilt training gamification \cite{stannett}. It is built with the multiplatform Electron framework. Most of the time, the app is designed to be invisible, only occupying a system tray slot. The app's GUI displays the live influence and momentum estimates. When using the win\% momentum, figures are rescaled to between -5 and 5. If tilt is detected or a motivational or explanatory `reality check' message is predicted to be useful, a notification is triggered before the next game, usually after entering the queue. A graph illustrates the player's live momentum and Elo statistics over time, also indicating wins and losses on the timeline, and past notifications from the app can also be reviewed and rated. Active learning is allowed by an optional but encouraged tilt survey, which uses time-stratified random sampling in order to maintain a naturalistic and enjoyable playing environment, and this will be a hidden setting post-beta, when enough data is collected. Notifications also deliver intelligence on break duration for optimal momentum.

\subsubsection{Adaptive Notification Strategy}

Initially, we begin with just the win\% momentum model, though our target is an intelligent agent that can accurately predict {\itshape felt} momentum and tilt, when to send a notification, and what the contents should be. The environment or the state at time $t$ (per second) is represented by the time-dependent momentum embedding, additional modalities and user activity (for example, mouse and keyboard press rate), the game client phase (post-game lobby, pre-game lobby or in-queue), and outputs from survey models. We propose a reinforcement learning approach to achieve this, with an action space corresponding to personal tilt trainer vocabulary, and optimal break duration. The reward function which the agent should maximise is proposed to be some combination of reduction in tilt prior to the next match played, increase in player skill, user satisfaction, or a user-defined objective \cite{christiano}. A phased, active, online learning rollout strategy is defined in order to maximise success, accuracy, applicability and computational performance of the platform.

\subsubsection{Longitudinal Survey}

A longitudinal survey may be ideal for assessing the final effectiveness of our system in long-term user satisfaction. In addition, we plan to test the hypothesis that, without hurting user interest, tilt management training may improve upon addiction score, as {\itshape League} is one of the most addictive games \cite{skarupova}, especially among youth \cite{bekir}, and self-regulation has been recommended as a shared conceptualisation tool, because neither `virtual life' nor real life suffer due to high self-regulation skills.



\subsection{Applications}\label{sec:otherAreas}

The methods presented in this paper are designed for gamers, however, the same methodology, with some adjustments, may easily be applied to other activities. For example, in skill-based gambling, whether these are part of an addictive behaviour that a gambler would like to minimise, recreational play, or professional efforts for which the user would like to minimise financial risk. This would be most useful to activities that involve the highest degree of tilt, strategic planning, long session times, and episodic event schedules where notification timing can be tapped. As with distinguishing a mistake from bad luck and deviance in the gaming case, the degree to which one has experienced misfortune can be modelled with relative ease, for example, by using the deviation of the player's profit or loss from the expected value (EV) of their actions. This may also characterise various other high pressure environments for which performance statistics and interaction patterns can be used to create beneficial notification strategies; i.e. financial trading, crisis management, emergency departments, and in sports. 

\pagebreak

\bibliographystyle{ACM-Reference-Format}
\bibliography{references}

\appendix

\section{Feature List}\label{sec:allfeats}


\begin{enumerate}
  \footnotesize \item \underline{Player ranked summary performance features} \\
  \footnotesize{elo} \hspace*{5px}-\hspace*{2px} \textit{Match average skill rating over all participants} \\
  \footnotesize{season win rate} \hspace*{5px}-\hspace*{5px} \textit{Player's win rate for the season} \\

  \item \underline{Base global champion stats} \\
  \hspace*{10px} \textit{Global stats for the chosen champion (in the current role and skill bracket)} \\
  \hspace*{10px} \textit{Stats for today, very recent trends in the meta:} \\
  \footnotesize{regional avg. champion [win rate, gold, creep score] {\itshape today}} \\
  \hspace*{10px} \textit{Stats for this patch (stable trends based on the game version):} \\
  \footnotesize{global avg. champion [win rate, play rate, kills, deaths, assists, kill sprees, gold, total damage taken, total heal, wards placed, wards killed, total damage, total magic damage, total physical damage, total true damage] {\itshape this patch}} \\
  \hspace*{10px} \textit{Since these are specific features for each role, the early, mid, and late game potential of the chosen champion is normalised by position:} \\
  \footnotesize{global avg. champion duration [0-15, 15-20, 20-25, ..., 40+] win rate {\itshape this patch}} \\
  
  \item \underline{Base global matchup stats} \\
  These are gathered for each role, and also for the four extra cross combinations that occur between the four players in the {\itshape Bottom} lane, which correspond to synergies and counters between {\itshape Marksman} and {\itshape Support} champions.
  
  \hspace*{10px} \textit{Global stats for the lane matchup (in the current role and skill bracket):} \\
  \footnotesize{global avg. matchup [wins, win rate, gold, creep score, total damage dealt to champions, champion.gg's `weighed score']} \\

  \item \underline{Player average season performance stats (history)} \\
  These features are averages across the player's champion pool, using data from the previous season too if there are not many games for the current season. As with the global statistics, because these are specific for each role, the performance for each category is normalised by the position.

  \hspace*{10px} \textit{Player diverse game knowledge/performance (avg. across champion pool):} \\
  \footnotesize{player champion average [wins, losses, kills, deaths, assists, gold, creep score, damage dealt, damage taken]} \\
  \hspace*{10px} \textit{Player performance over last 2 seasons for specific champion categories:} \\
  \footnotesize{player champion class [Fighter, Tank, Mage, Assassin, Support, Marksman] average [wins, losses]} \\

  \item \underline{Player champion-specific performance stats (proficiency/playstyle)} \\
  \hspace*{10px} \textit{Player recent game performance on their chosen champion (average for last 2 seasons). Since this feature is not normalised for the champion, this is essentially their predicted performance stats for this game:} \\
  \footnotesize{champion proficiency [games, wins, losses, bayes win rate, alltotal wins, alltotal losses, alltotal bayes win rate, kills, deaths, assists, gold, creep score, damage taken]} \\
  \hspace*{10px} \textit{The previous group of features, normalised by the global avg. (representing player skill on the champion they've chosen):} \\
  \footnotesize{champion proficiency kills $\div$ global avg. champion kills {\itshape this patch}} \\
  \footnotesize{champion proficiency deaths $\div$ global avg. champion deaths {\itshape this patch}} \\
  \footnotesize{champion proficiency assists $\div$ global avg. champion assists {\itshape this patch}} \\
  \footnotesize{champion proficiency gold $\div$ regional avg. champion gold {\itshape today}} \\
  \footnotesize{champion proficiency creep score $\div$ regional avg. champion creep score {\itshape today}} \\
  \footnotesize{champion proficiency damage taken $\div$ global avg. champion damage taken {\itshape this patch}} \\

  \item \underline{Player momentum/tilt performance stats (of the most recent games)} \\
  \footnotesize{recent duration} - \textit{How long the recent match lasted (in general, longer is better - more carrying losing teams, less solo losing games)}\\
  \footnotesize{recent time since match} - \textit{Time since match occurred}\\
  \footnotesize{recent time of day} - \textit{Approximate time of day when match occurred (hours since midnight, for the given region)}\\
  \footnotesize{recent champion classes}\footnote[1]{Only included for neural network models (not selected in SFFS for logistic regression).} - \textit{Class membership of the chosen champion in match}\\
  \footnotesize{recent [win rate, kills, deaths, assists, creep score, kill participation, control wards bought] - \textit{Performance stats for recent games}} \\
  \hspace*{10px} \textit{How skilled this player is at the champions they've been playing very recently:} \\
  \footnotesize{recent champion proficiency [games, wins, losses, bayes win rate]} \\
  \hspace*{10px} \textit{How good the champions that this player is playing are in the current meta:} \\
  \footnotesize{recent regional avg. champion win rate {\itshape today}} \\
  \footnotesize{recent global avg. champion win rate {\itshape this patch}} \\
  \hspace*{10px} \textit{How well this player has been playing the champions they've been playing compared to how well they normally play them}\textsuperscript{(momentum)}{\itshape :} \\
  \footnotesize{[recent kills, deaths, assists, creep score $-$ champion proficiency kills, deaths, assists, creep score]} \\

\end{enumerate}









\end{document}